\documentclass[final]{cvpr}

\usepackage{times}
\usepackage{epsfig}
\usepackage{graphicx}
\usepackage{amsmath}
\usepackage{amssymb}
\usepackage{subcaption}
\usepackage{multirow}

\usepackage[pagebackref=true,breaklinks=true,colorlinks,bookmarks=false]{hyperref}

\def\cvprPaperID{8} 
\def\confYear{CVPR 2021}

\begin{document}

\title{Directional GAN: A Novel Conditioning Strategy for Generative Networks}

\author{Shradha Agrawal\\
Adobe \\
{\tt\small shradagr@adobe.com}
\and
Shankar Venkitachalam\\
Adobe\\
{\tt\small svenkita@adobe.com}

\and
Dhanya Raghu\thanks{This work was done during author's internship with Adobe.}\\
Amazon\\
{\tt\small dhanyr@amazon.com}

\and
Deepak Pai\\
Adobe\\
{\tt\small dpai@adobe.com}

}

\maketitle

\begin{abstract}
   Image content is a predominant factor in marketing campaigns, websites and banners. Today, marketers and designers spend considerable time and money in generating such professional quality content. We take a step towards simplifying this process using Generative Adversarial Networks (GANs). We propose a simple and novel conditioning strategy which allows generation of images conditioned on given semantic attributes using a generator trained for an unconditional image generation task. Our approach is based on modifying latent vectors, using directional vectors of relevant semantic attributes in latent space. Our method is designed to work with both discrete (binary and multi-class) and continuous image attributes. We show the applicability of our proposed approach, named Directional GAN, on multiple public datasets, with an average accuracy of 86.4\% across different attributes.
\end{abstract}

\section{Introduction}
\label{sec:introduction}

Image content is a predominant factor in marketing campaigns (such as Ads, emails), websites, banners among others. Marketers typically rely on content designers to create such content. Content designers in-turn rely on tools such as Adobe Photoshop to edit, polish and composite the captured raw images to create appropriate content. Marketers and Designers iterate back and forth during such content creation process. It would be ideal for the marketers or designers, to be able to specify their requirements to an automated algorithm, which can generate content for them instantly. If the user is not satisfied with the result, they can simply discard the generated image.  In addition, it is non-trivial to create minor variations of content, personalized to different market segments. We aim to simplify this process by leveraging GAN and a novel conditioning strategy.


Recent years have witnessed exponential growth of Artificial Intelligence, Neural Networks and specifically GANs. Once trained, they can generate an entirely new image similar in nature to the training data. Although GANs typically generate low resolution images, recent work \cite{karras2017progressive,karras2019style} has shown that it is feasible to generate good quality, high resolution images. However, these networks generate images from latent vectors that are randomly sampled from the latent space and hence unable to control the semantic attributes of the images. Conditional GANs \cite{mirza2014conditional, lu2018attribute} provide the user the ability to customize the desired attributes. However, this requires re-training of the network with conditional adversarial loss, for the choice of attributes. Further, addition of new attributes would require re-training and re-designing the network. Alternatively, one can use style transfer \cite{huang2017arbitrary} to achieve the desired results but that requires an image with desired attributes. 

Radford \etal \cite{radford2015unsupervised} have shown that linear interpolation between two random latent vectors $\boldsymbol{z_1}$ and $\boldsymbol{z_2}$ results in smooth transition between generated images. We leverage this observation by hypothesising that it is feasible to learn hyperplanes that separate distinct image attribute values. We could then move the latent vector in the appropriate direction such that it can generate an image with the desired attributes, thus achieving the conditioning effect. However, if various attributes of the image are entangled in latent space, then changes in one attribute might result in unintended changes in other attributes. Recently, StyleGAN \cite{karras2019style} has shown that image features in the latent space are sufficiently disentangled, allowing us to change only the desired attributes, without affecting others.

In this work, we propose Directional GAN (DGAN), a simple and novel approach to generate high resolution images conditioned on desired semantic attributes. Our approach simplifies the complex task of conditional generation by separating the process of generation from that of conditioning and allows them to be independent. We start with a generator (GAN) that generates a realistic image from a random latent vector and propose a mechanism to move this latent vector appropriately in the latent space to achieve the effect of conditioning in the generated image. We do this by learning linear hyperplanes in the latent space that separate the attribute values (say tee from dress).

We validate our approach in two ways, 

1. By generating full-body human images with the ability to condition on the style of clothing (tee or dress) and the pose (front and back) on MPV \cite{dong2019towards} and Deepfashion \cite{liu2016deepfashion}.

2. By providing the ability to condition on facial attributes such as hair-color (black, brown or blonde), gender and degree of smile on CelebA-HQ face dataset.

Our contribution in this work is multi-fold. First, we propose an approach employing directional vectors to allow for conditioning in GANs. We show mathematically, that using our approach, we can move the latent vector to the desired subspace in a single step. Next, we show the applicability of our method not only for single attribute conditioning but also multiple attributes together. Our approach works with discrete (binary and multi-class) as well as continuous space of attribute values. Finally, our approach maintains the same Frechet Inception Distance (FID) \cite{heusel2017gans} score as that of unconditional generation, 23 for Full-body Dataset and 5.06 for CelebA-HQ. Hence allows for conditioning without deteriorating the quality of generated images, unlike in \cite{yildirim2019generating} where conditional generation of images achieves a worse FID score (9.63) than that of unconditional generation (5.15) . Although we demonstrate our approach with StyleGAN, it is generic enough to be applicable on any GAN with sufficiently disentangled image features in latent space.

The rest of the paper is organized as follows: Section \ref{Related Work} discusses recent relevant literature in this area and their drawbacks. Our method is detailed in Section \ref{Approach}. Experimental details starting with characteristics of datasets, pre-processing steps and results are presented in Section \ref{Experiments}.

\begin{figure*}
\begin{center}
\includegraphics[scale=0.31]{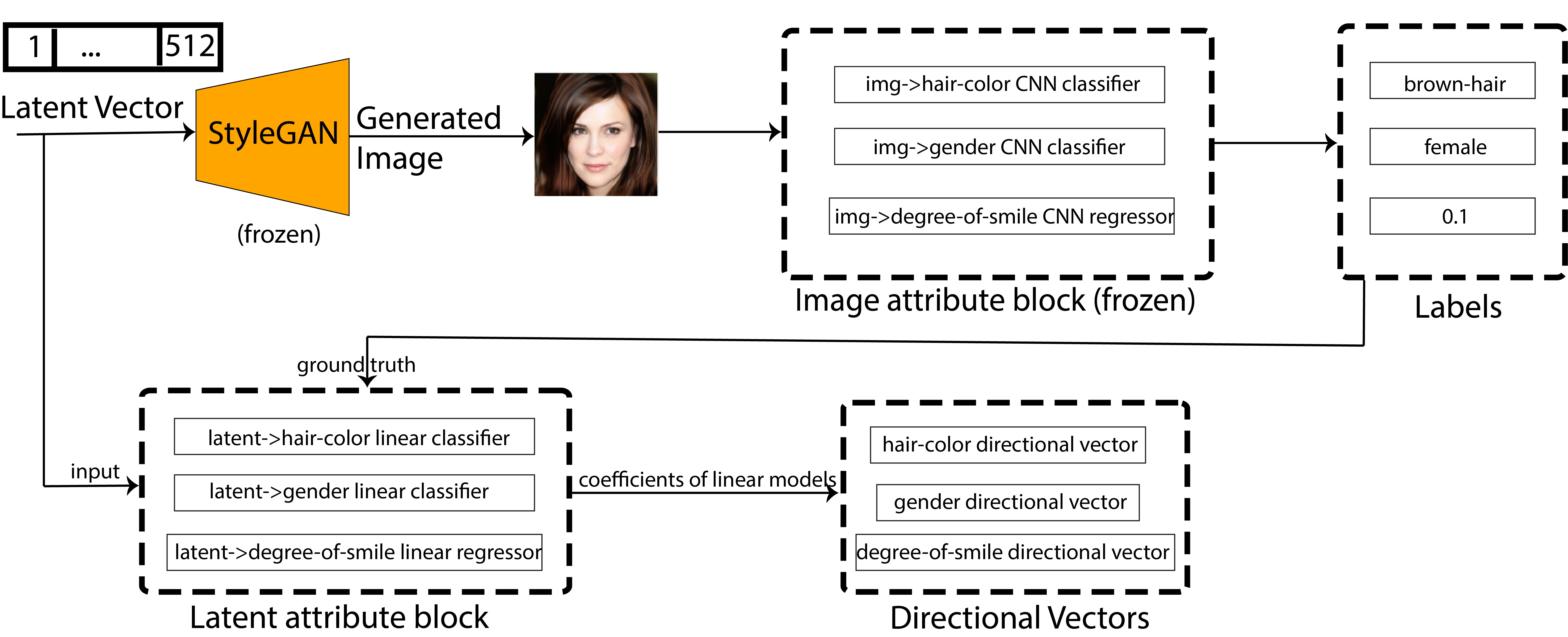}
\end{center}

    \caption{Block diagram depicting the training of classifiers/regressors in the latent attribute block. The GAN (frozen parameters) generates an image for a given latent vector. Classifiers/regressors (frozen parameters) in the \textit{image-attribute} block predict labels for various attributes of this image. These labels, along with the given latent vector, are used to train classifiers/regressors in the \textit{latent-attribute} block to learn separating hyperplanes/regression lines in the latent space. These hyperplanes provide the directional vectors which are used in conditional image generation.}
\label{fig:approach}
\end{figure*}

\section{Related Work}
\label{Related Work}
Generative models for content creation today are powered by deep neural networks, specifically \textit{Variational Auto Encoders} (VAEs) \cite{kingma2013auto} and \textit{Generative Adversarial Networks} (GANs) \cite{goodfellow2014generative}. GANs are generally preferred over VAEs since they tend to generate sharper images. However, in both networks, images are generated from randomly sampled latent variables and hence offer no manual control in the generation process. For instance, GANs proposed in \cite{karras2017progressive} can generate images of human faces, but do not allow for the control over facial attributes such as color of skin, eye or hair among others. Conditional GAN (cGAN) \cite{mirza2014conditional} attempts to address this issue by providing the ability to control certain aspects of content generation. Lu et al. \cite{lu2018attribute}, for instance, have demonstrated the generation of facial images conditioned on attributes such as hairstyle, glasses, gender, smile among others. However, introduction of new attributes would require re-training cGANs and possibly re-designing the network. 
Yildirim et al. \cite{yildirim2019generating} proposes to use embeddings with the latent vectors to allow for conditioning in StyleGAN. However, their FID score deteriorates from it's unconditional sibling.

Alternatively, conditioning could be achieved with style transfer \cite{huang2017arbitrary}, where one starts with a randomly generated image from a GAN and uses characteristics of a different image for applying the necessary conditioning. This approach, also referred to as style mixing, is applied by StyleGAN\cite{karras2019style} to generate human faces. However, using such an approach will require numerous runs of the generator to generate the image with the required characteristics, using an architecture such as StyleGAN. Since the number of generator runs that might be required is unbounded, the technique is inefficient and compute intensive.

Our approach, on the other hand, utilizes an unconditional GAN, and achieves conditioning by learning to manipulate the latent vector input to this GAN. This area of research is fairly recent and only very few papers have been published, the most notable ones being \cite{shen2020interpreting} and \cite{abdal2020styleflow}. While these methods require multiple steps of gradually changing the attributes to generate an image with the required attributes, our method can achieve the same in a single step. In addition, as opposed to the existing works, our approach is also able to condition on multi-class attributes.

\section{Approach}
\label{Approach}
We propose an approach called Directional GAN (DGAN) which allows for control over the latent input, using directional vectors in the latent space. We assume, and later prove empirically that, by learning a linear hyperplane that separates the feature values in latent space, we can achieve desired conditioning by forcing the latent vector to be in the desired subspace. We do so by learning two sets of classifiers, one set that predicts attributes given an image and a second set that predicts attributes given a latent vector.  


Our architecture has three components, a GAN which generates realistic images from random vectors, the \textit{image-attribute} block and the \textit{latent-attribute} block. The generator and discriminator follow the network architecture proposed in StyleGAN \cite{karras2019style}. The \textit{image-attribute} block identifies the attribute labels in the generated images, which are then used by the \textit{latent-attribute} block to learn separating hyperplanes or regression lines in the latent space.

The \textit{image-attribute} block consists of multiple classifiers/regressors, one per conditioning attribute. For each discrete valued attribute, the \textit{image-attribute} block consists of a classifier whose task is to take the image generated by the GAN and determine an appropriate label for the image. For each continuous valued attribute, the image attribute block consists of a regressor which determines an appropriate real value for the image. For instance, one classifier determines if the clothing in generated image is tee or dress. Similarly, a regressor determines the degree of smile from the generated image. These labels are then associated with the latent vector which generated the image. The latent vectors and their corresponding labels form the input to the second block, the \textit{latent-attribute} block.

The \textit{latent-attribute} block consists of one classifier/regressor per conditioning attribute. In case of multi-valued discrete attribute, one could learn as many one-vs-all classifiers as the number of possible values of the attribute. Alternatively, methods such as multinomial logistic regression could be used, eliminating the need for learning multiple classifiers separately. Each of these classifiers/regressors take as input the latent vector and the attribute label to learn a linear boundary separating the attribute types, like tee/dress or regression line for continuous features like degree of smile, in the latent space. We then use the \textit{latent-attribute} block to move the latent vector to the latent subspace with desired attributes for conditioning. 
Section \ref{Training} outlines our training process and Section \ref{sec:image_generation} describes image generation conditioned on given attributes, after completion of training. 



\begin{figure*}
    \centering
    \includegraphics[width=\linewidth]{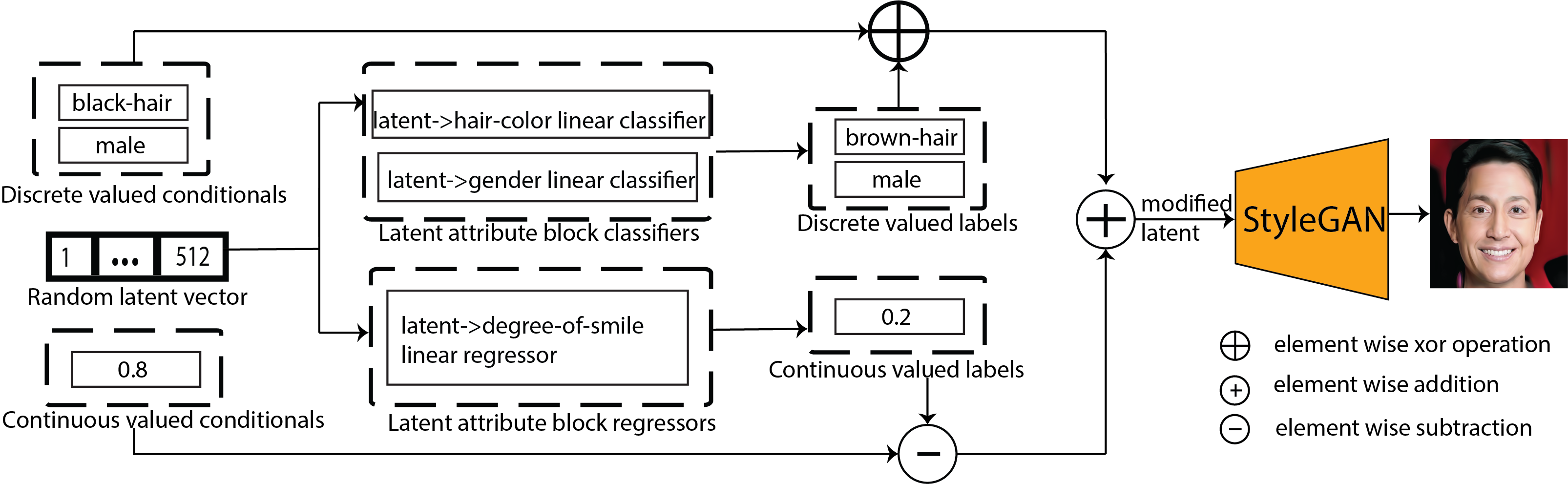}
    \caption{Our image generation process starts with randomly sampling a latent vector and moving it to the desired subspace of attributes. The latent vector is modified with respect to each attribute where value of latent vector doesn't match with input conditioning vector.}
    \label{fig:image-generation}
\end{figure*}

\subsection{Training}
\label{Training}
One of the major advantages of our architecture design is the modular training of its three components. The first two components, the GAN and the \textit{image-attribute} block are trained independently from each other. Training of the third component, the \textit{latent-attribute} block, requires outputs from the first two components whose parameters are kept frozen (See Figure \ref{fig:approach}).  


Each classifier/regressor in the \textit{image-attribute} block is trained to accurately label one attribute class or a real value. For instance, we would train one classifier and one regressor, first to predict the clothing style, tee or dress, and second to predict the degree of smile. Images generated from GAN are fed into the \textit{image-attribute} block. The outcome of this block is a set of labels, one per attribute class.  

Each classifier/regressor in the \textit{latent-attribute} block is trained to learn a linear separating hyperplane in the latent space that separates the classes of a given attribute, or a regression line for continuous valued features. In the above example, we train one linear classifier and one regressor, first for clothing style and second for degree of smile. While the latent vector forms the input feature space, the outcome generated by the \textit{image-attribute} block provides the ground truth labels. The learned coefficients from this block form our directional vectors.

Let,
\begin{equation}
    \mathcal{D} = \{X_i\}_{i=1}^{N}
\end{equation}
be the dataset with $N$ images, $X_1,\dots,X_N$, and let $G$ denote the generator in our GAN. 
\begin{equation}
    G: \boldsymbol{z} \xrightarrow{} \tilde{X} 
\end{equation}
where $\boldsymbol{z}$ denotes the latent vector and $\tilde{X}$ denotes the generated image. Also, let $\boldsymbol{p_d}$ denote the conditional vector for discrete attributes and $\boldsymbol{p_r}$ for continuous attributes.
\begin{equation}
    \boldsymbol{p_d}:  [cd_1, cd_2, ..., cd_n]^T
\end{equation}
where $cd_i$ is the desired class of $i^{th}$ discrete valued attribute.
\begin{equation}
    \boldsymbol{p_r}:  [y'_1, y'_2, ..., y'_m]^T
\end{equation}
where $y'_i$ is the desired value of $i^{th}$ continuous valued attribute.
\subsubsection{For discrete valued attributes:}

Let $C_i^{IA}$ be the classifier of \textit{image-attribute} block for $i^{th}$ attribute and $C_i^{LA}$ be the linear classifier of \textit{latent-attribute} block for $i^{th}$ attribute. 
\begin{equation}
    C_i^{IA} : \tilde{X} \xrightarrow{} \textrm{class of $i^{th}$ attribute}
\end{equation}
\begin{equation}
    C_i^{LA} : \boldsymbol{z}\xrightarrow{} \boldsymbol{d_i} \cdot \boldsymbol{z} + b
\end{equation}
where $\boldsymbol{d_i}$ is directional vector for $i^{th}$ attribute.

\subsubsection{For continuous valued attributes:}

Let $R_i^{IA}$ be the regressor of \textit{image-attribute} block for $i^{th}$ attribute and $R_i^{LA}$ be the linear regressor of \textit{latent-attribute} block for $i^{th}$ attribute. 
\begin{equation}
    R_i^{IA} : \tilde{X} \xrightarrow{} \textrm{real value of $i^{th}$ attribute}
\end{equation}
\begin{equation}
    R_i^{LA} : \boldsymbol{z}\xrightarrow{} \boldsymbol{d_i} \cdot \boldsymbol{z} + b
\end{equation}
where $\boldsymbol{d_i}$ is directional vector for $i^{th}$ attribute.

\subsection{Image Generation}
\label{sec:image_generation}
The input to the image generation process (See Figure \ref{fig:image-generation}) starts with a vector of attribute specifications. Each element of the vector is an indicator variable, for discrete attributes, or a real value that specifies the label of the desired attribute in the generated image. We start with a randomly generated latent vector, which is passed through the latent attribute block. The classifiers/regressors in this block, output the labels corresponding to each attribute. Discrete valued attributes are one hot encoded and thus the output is \textit{XOR}ed with the conditioning vector from the input, which results in a vector \textbf{c} with \textit{1} where the latent vector is likely to generate an attribute different from the desired outcome and \textit{0} otherwise. For the attributes that need nudging, to generate the desired outcome, we move the latent vector in the appropriate direction along the linear combination of directional vectors. For each such attribute, we first compute the signed distance of the latent vector from the separating hyperplane. We then move the latent vector in the required direction, by an amount equal to the signed distance and some $\delta$, since signed distance will place the point exactly on the hyperplane. For continuous valued attributes, if the desired value of the attribute is different from the current value that the latent vector has, the latent vector is moved in the direction specified by the direction vector by a distance equal to the difference.

\subsubsection{For discrete valued attributes:}

Let, 
\begin{equation}
     H_i : \boldsymbol{d_i} \cdot \boldsymbol{x} + b = 0
\end{equation}
be the equation of the hyperplane for the $i^{th}$ latent attribute classifier. Here $\boldsymbol{d_i}$ is the direction vector and $b$ is intercept. Let, 
\begin{equation}
    D_c = 
\begin{bmatrix}
    \boldsymbol{\hat{d_1}}, \boldsymbol{\hat{d_2}}, ..., \boldsymbol{\hat{d_n}} \\
\end{bmatrix}^T
\end{equation}

be the matrix with $i^{th}$ row denoting unit direction vector $\boldsymbol{\hat{d_i}}$ for $i^{th}$ discrete valued attribute.
Let,
\begin{equation}
    \boldsymbol{q_d} = [\textrm{class}(\boldsymbol{d_i} \cdot \boldsymbol{z} + b)]_{i=1}^{n}
\end{equation}
be the vector with $i^{th}$ element denoting class of $i^{th}$ discrete attribute for $\boldsymbol{z}$.

We update,
\begin{equation}
    \underset{1\times 512}{\boldsymbol{z'}} = \underset{1\times 512}{\boldsymbol{z}} - [\underset{1\times n}{\boldsymbol{c}} \circ (\underset{1\times n}{\boldsymbol{s}} + \delta )] \times \underset{n\times 512}{D_c}
\end{equation}

where $\boldsymbol{c}$ is the choose vector, output of xor operation
between input conditioning vector and latent attribute vector and $\boldsymbol{s}$ is vector of signed distances where each $s_i$ is the  signed distance from $\boldsymbol{z}$ to hyperplane $H_i$ given by,
\begin{equation}
    s_i = -\frac{ \boldsymbol{d_i} \cdot  \boldsymbol{z} + b}{||\boldsymbol{d_i}||}
\end{equation}
And $\boldsymbol{c}$ is defined as,
\begin{equation}
    \boldsymbol{c} = \boldsymbol{p_d} \oplus \boldsymbol{q_d}
\end{equation}

\subsubsection{For continuous valued attributes:}

Let, 
\begin{equation}
     D_r = 
\begin{bmatrix}
    \boldsymbol{\hat{d_1}}, \boldsymbol{\hat{d_2}}, ..., \boldsymbol{\hat{d_m}} \\
\end{bmatrix}^T
\end{equation}

be the matrix with $i^{th}$ row denoting unit direction vector $\boldsymbol{\hat{d_i}}$ for $i^{th}$ continuous valued attribute.
And,
\begin{equation}
    \boldsymbol{q_r} = [y_1, y_2, ..., y_m]^T
\end{equation}

be the vector with $i^{th}$ element denoting value of $i^{th}$ continuous valued attribute for $\boldsymbol{z}$.

Define, 

\begin{equation}
    \Delta = \boldsymbol{p_r} - \boldsymbol{q_r}
\end{equation}


\begin{equation}
    \underset{1\times 512}{\boldsymbol{z'}} = \underset{1\times 512}{\boldsymbol{z}} + 
    \underset{1\times m}{\Delta} \times \underset{m\times 512}{D_r}
\end{equation}

\subsubsection{Combining the updates:}
Combining the updates into a single equation:
\begin{equation}
    \label{eq:final_update}
    \underset{1\times 512}{\boldsymbol{z'}} = \underset{1\times 512}{\boldsymbol{z}} - [\underset{1\times n}{\boldsymbol{c}} \circ (\underset{1\times n}{\boldsymbol{s}} + \delta )] \times \underset{n\times 512}{D_c} + \underset{1\times m}{\Delta} \times \underset{m\times 512}{D_r}
\end{equation}


\section{Experiments}
\label{Experiments}
In this section we describe the details of our experiments, the datasets used, pre-processing steps, machine configuration and training setup among others.

\subsection{Datasets}
\label{Datasets}
\subsubsection{Full-body Dataset}
We use the union of two public datasets for training our generator and discriminator- Multi Pose Virtual Try On (MPV) \cite{dong2019towards} and Deep Fashion (DF) landmark detection benchmark \cite{liu2016deepfashion}. Images in both datasets have variation in pose and clothing. Additionally, we use Deep Fashion (DF) category attribute prediction dataset \cite{liu2016deepfashion} for training our \textit{image-attribute} block.
Table \ref{tab:Characteristics_datasets} summarizes the characteristics of our datasets. Further, Deep fashion datasets contains annotations for clothing type. We choose images from 2 categories tee and dress, among the 50 categories of available images. 

We use this dataset to demonstrate the generation of high resolution full body images conditioned on binary attributes, clothing style (tee, dress) and pose (front, back).

\subsubsection{CelebA-HQ}
We apply the proposed DGAN on the CelebA-HQ dataset \cite{karras2017progressive} and demonstrate generation of high resolution face images conditioned on hair-color (black, brown and blonde), gender (female and male) and degree of smile([0, 1]). While gender and hair-color are discrete attributes with hair-color being multi valued, smile is a continuous valued attribute with values in $[0, 1]$. 

\subsection{Data Pre-processing}
\label{Data Pre-processing}
The MPV and DF datasets require pre-processing for our task of generating full body human images. For instance, the dataset contains half-body images as well as images with only clothes. We first filter the data, to retain only full body human images using available annotations. However, we noticed discrepancies with annotations in deep fashion where images up-to knee length, were marked as full-body. This is likely because the clothing, which is the dominant part of the dataset, was complete. We also noticed that images with only clothing, but no human, were marked as full-body. To further cleanse our dataset and only retain full-body human images, we used tensorflow object detection API with SSD \cite{liu2016ssd} mobile net \cite{howard2017mobilenets} trained on COCO dataset \cite{lin2014microsoft}. After cropping, we have about 21,000 usable images. However, these images were of varying resolutions. Training GAN requires the input images to be of the same size, so that the architecture, including GPU processing could be optimized. Hence, we re-scale our images to be of constant 128x512 resolution. We then apply SR-GANs \cite{ledig2017photo} to reduce artifacts introduced in re-scaling operation. 

No pre-processing steps were performed for CelebA-HQ dataset.

\begin{table*}
\begin{center}
\caption{Characteristics of datasets}
\label{tab:Characteristics_datasets}
\begin{tabular}{c|c|c|c}
\hline
Name & Resolution & N(images) & N(images) after preprocessing\\
\hline
CelebA-HQ & 1024x1024 & 30,000 & 30,000\\
MPV & 256x192 & 35,687 & 13713\\
DF landmark detection & 200x200 to 512x512 & 123,016 & 6993\\
\hline
\end{tabular}
\end{center}
\end{table*}

\subsection{Training Details}
\label{Training Details}
\subsubsection{Full-body dataset}
We build upon the official Tensorflow \cite{abadi2016tensorflow} implementation of StyleGAN \cite{karras2019style} from which we inherit most of the training details. Our training time is approximately one week on 1 Nvidia Tesla K80 GPU for 300 epochs. 

We modify the network architecture proposed in StyleGAN \cite{karras2019style} to allow for rectangular input resolution.
The resolution of layers starts with 2x8 and progresses till 128x512 in increments of 2. We test Directional GAN with two attributes- pose and clothing category. For the \textit{image-attribute} block, we use pre-trained classifiers for each of these attributes. We use a resnet-50 \cite{he2016deep} model pre-trained on Deep Fashion category attribute prediction dataset \cite{liu2016deepfashion} to classify clothing style of a generated image into two classes - tee, dress; and open pose \cite{cao2018openpose} to classify the pose into two classes - front and back. 
The \textit{latent-attribute} block consists of two linear classifiers which learn the separating hyperplanes for clothing category and pose. We then generate images with desired attributes using the procedure outlined in section \ref{sec:image_generation}.
\subsubsection{CelebA-HQ}
We used the pre-trained generator from the official StyleGAN \cite{karras2019style} repository. Our \textit{image-attribute} block uses Microsoft Face API \cite{microsoft-face} for gender categorization into male or female, hair-color categorization into black, brown or blonde and prediction of degree of smile. Now, the latent attribute block learns a binary classifier for gender, a multi-class classifier for hair-color (which was achieved in a single pass using multinomial logistic regression) and a regression line for degree of smile. We then generate images with desired attributes using the procedure outlined in section \ref{sec:image_generation}.

\subsection{Evaluation}
To evaluate the end-to-end workflow of DGAN, we randomly sample latent vectors. For each of the sampled latent vector, we randomly generate a conditioning vector. We then use DGAN to get the resultant latent vector as well as the generated image. We run our classifiers/regressors from the \textit{latent-attribute} block to evaluate how often the latent vector falls in the desired subspace. We run our classifiers/regressors in the \textit{image-attribute} block on the generated image and compare how accurately the predicted attribute values match with the attribute values in the conditioning vector.

\begin{figure}
\centering
\begin{tabular}{cccc}
     \includegraphics[width=0.22\linewidth, keepaspectratio]{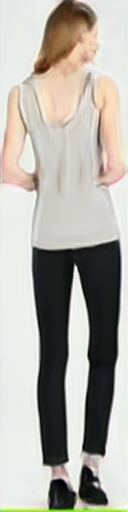}
     &
     \includegraphics[width=0.22\linewidth, keepaspectratio]{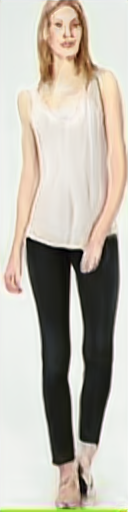}
     &
     \includegraphics[width=0.22\linewidth, keepaspectratio]{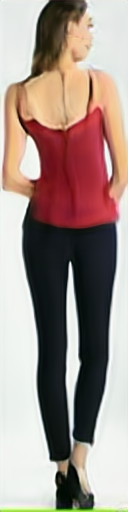} 
     &
     \includegraphics[width=0.22\linewidth, keepaspectratio]{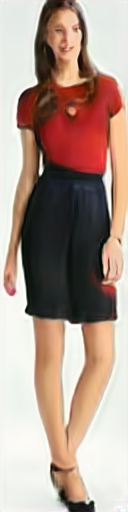} \\
     (a) & (b) & (c) & (d)
\end{tabular}
\caption{Conditioning on full body dataset. (a) depicts an image generated by a randomly sampled latent vector that happens to be in the subspace of back pose. (b) depicts the image generated by moving the latent vector of (a) to the subspace of front pose. (c) depicts an image generated by a randomly sampled latent vector that happens to be in the subspace of back pose and tee. (d) depicts the image generated by moving the latent vector of (c) to the subspace of front pose and dress.}
\label{fig:df}
\end{figure}

\section{Results}
\label{Results}

\subsection{Performance metrics}
We use an evaluation scheme on similar lines as that of \cite{park2019semantic}. For discrete attributes, we compute the accuracy, and for continuous valued attributes, we compute Root Mean Square Error (RMSE).  We compute these metrics not only for the conditional image generation but also for the latent vector manipulation. We also report FID and inception scores for capturing quality of our generated images in the full-body dataset.

\begin{figure}
\centering
\begin{tabular}{c}
     \includegraphics[width=0.7\linewidth, keepaspectratio]{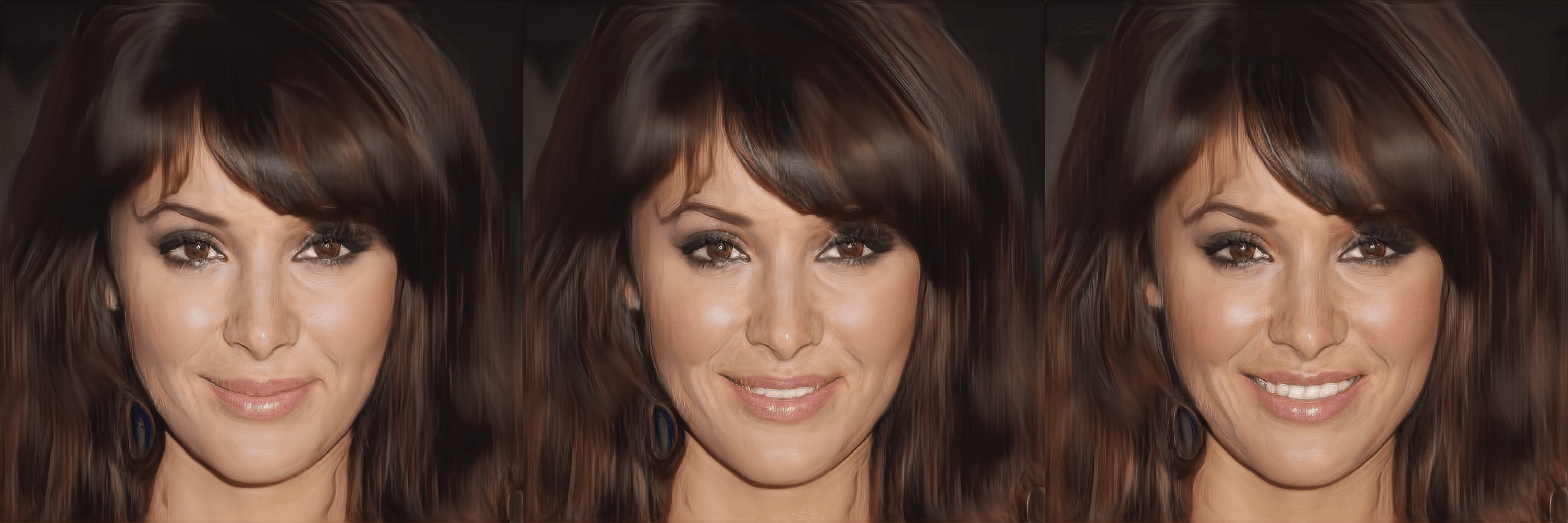}
     \\
     \includegraphics[width=0.7\linewidth, keepaspectratio]{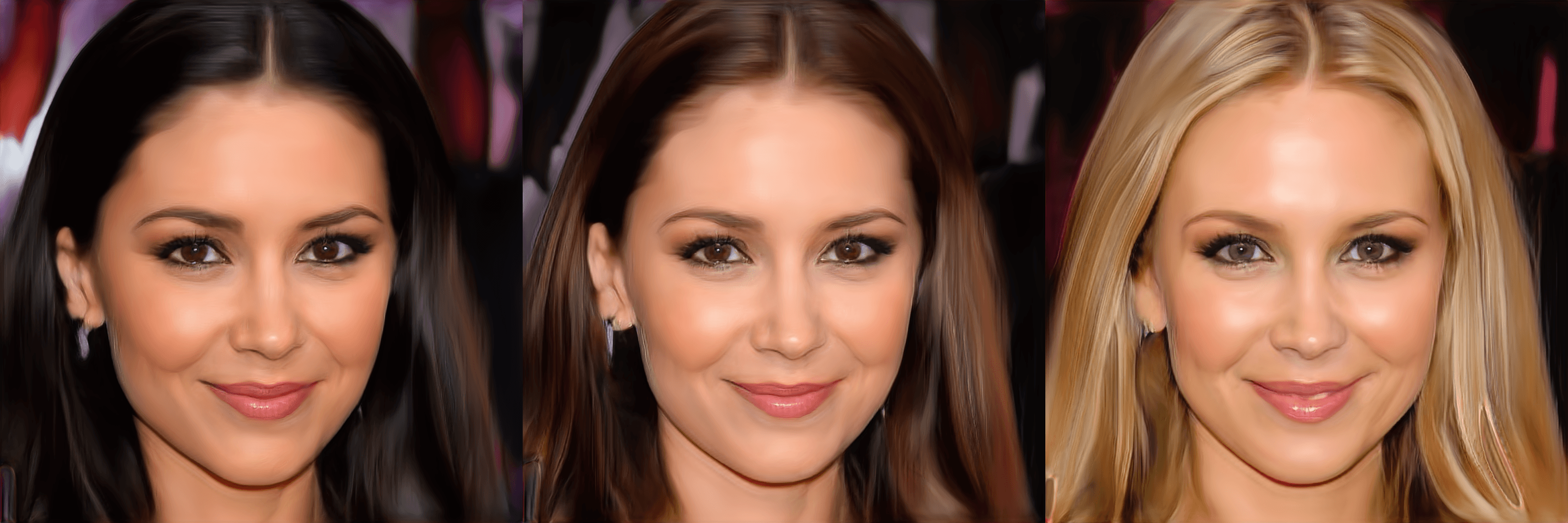}
     \\
     \includegraphics[width=0.7\linewidth, keepaspectratio]{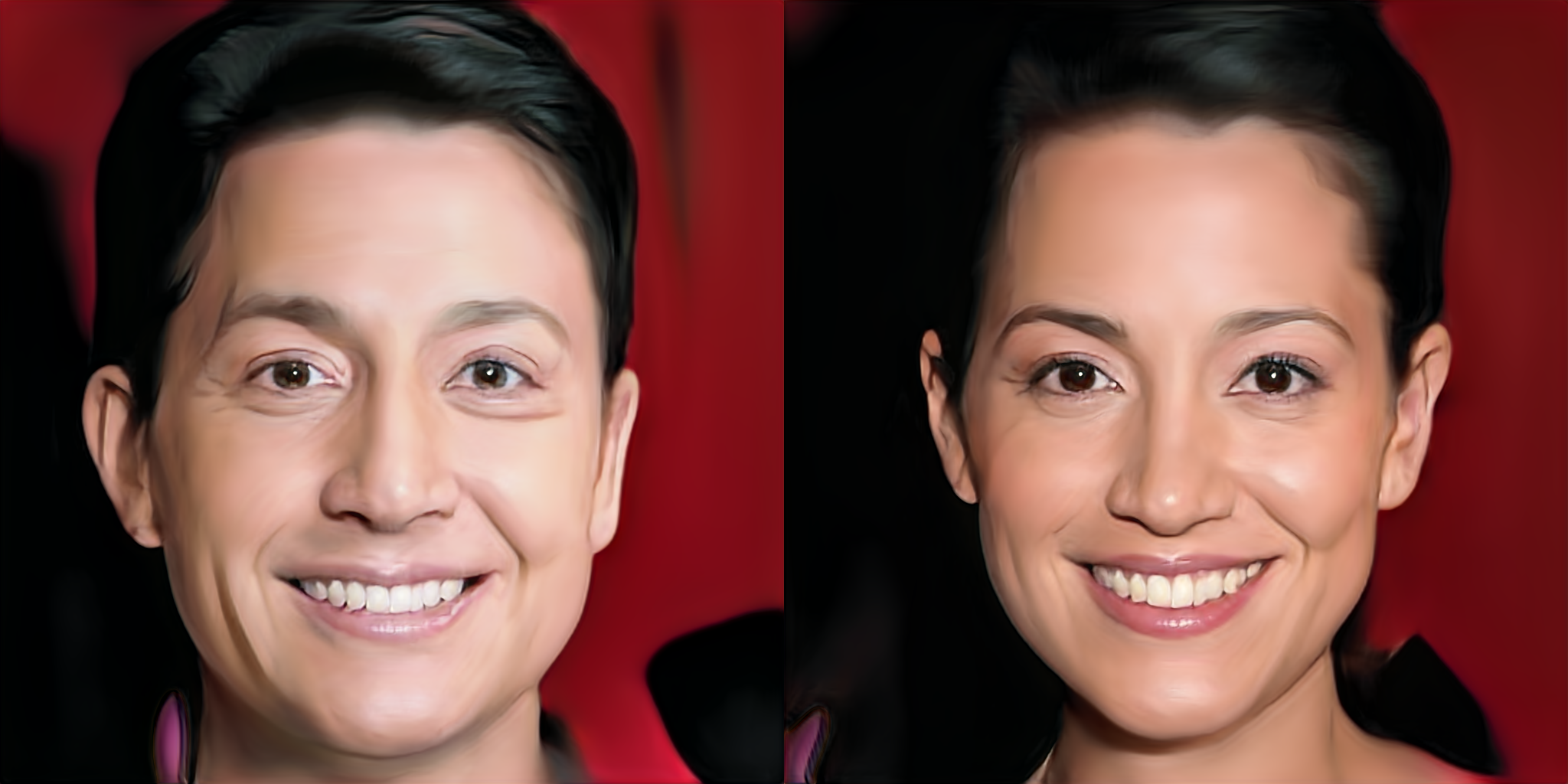} 
\end{tabular}
\caption{Conditioning on single attribute on celebA-HQ. From left to right on each row, $1^{st}$ row shows conditioning on increasing degree of smile, $2^{nd}$ row shows conditioning on hair-color from black to brown to blonde while $3^{rd}$ row shows conditioning on gender from male to female. }
\label{fig:celebA-single}
\end{figure}

\begin{figure}
\centering
\begin{tabular}{c}
     \includegraphics[width=0.7\linewidth, keepaspectratio]{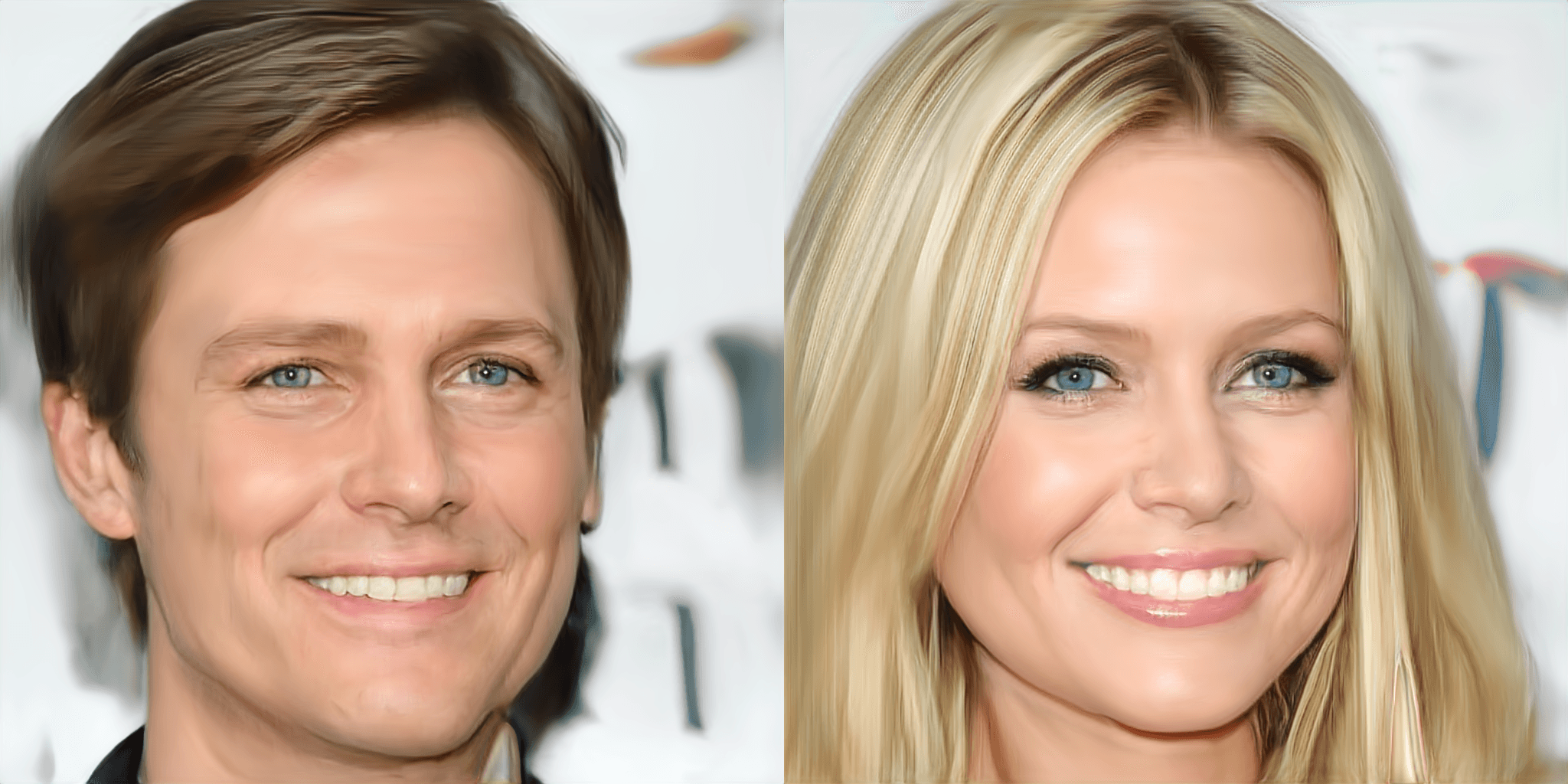}
     \\
     \includegraphics[width=0.7\linewidth, keepaspectratio]{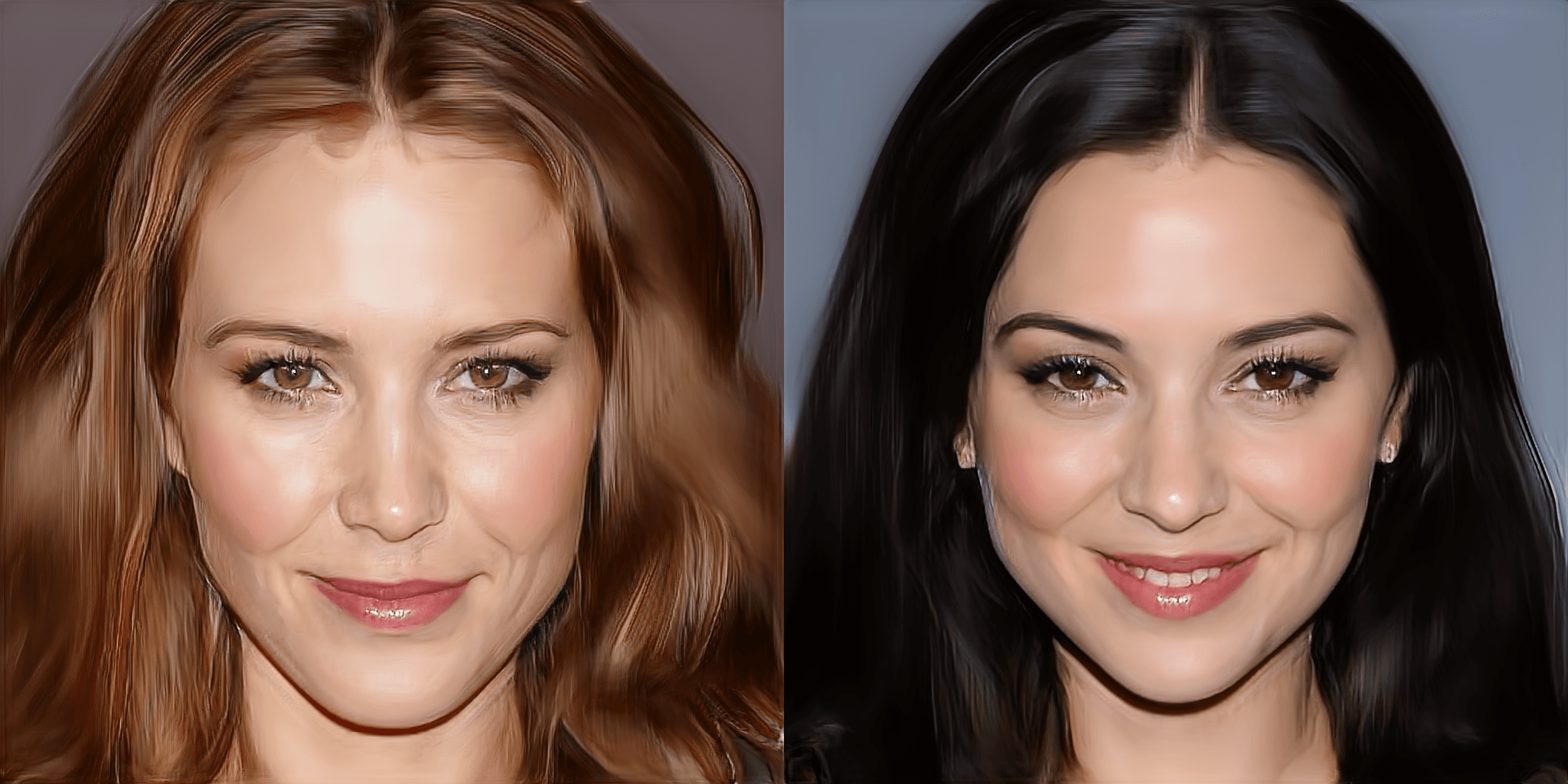}
   
\end{tabular}
\caption{Conditioning on multiple attributes on celebA-HQ. From left to right on each row, $1^{st}$ row shows conditioning on gender and hair-color from black haired male to blonde haired female, $2^{nd}$ row shows conditioning on degree of smile and hair-color from not smiling brown haired female to smiling black haired female. }
\label{fig:celebA-multi}
\end{figure}

\subsection{Full-body Dataset}
Our generated images achieve an FID score of 23. In table \ref{tab:DF_IS}, we have compared the inception score of our generated full-body images with the state-of-the-art. Our proposed approach (DGAN) outperforms state-of-the-art. 
In Figure \ref{fig:df}, (b) shows the image generated conditioned on front pose starting with latent vector of (a) while (d) shows the generated image after conditioning on both pose and clothing category starting with latent vector of (c). 
As shown in Table \ref{tab:DGAN_accuracy}, we achieve over 90\% accuracy in conditioning on clothing style and over 87\% accuracy in conditioning on pose.

\subsection{CelebA-HQ}
Since we directly use the pre-trained generator from StyleGAN \cite{karras2019style}, our generated images achieve the same FID score of 5.06 as noted in StyleGAN \cite{karras2019style}. Figure \ref{fig:celebA-single} shows the images generated, conditioned on a single attribute at a time with $1^{st}$ row showing conditioning on degree of smile, $2^{nd}$ row on hair-color and $3^{rd}$ row on gender.
Figure \ref{fig:celebA-multi} shows the image generated, conditioned on multiple attributes at a time with $1^{st}$ row showing conditioning on gender and hair-color and $2^{nd}$ row on degree of smile and hair-color. 
As shown in Table \ref{tab:DGAN_accuracy}, we achieve over 89\% accuracy in conditioning on gender and over 78\% accuracy in conditioning on hair-color and a RMSE of 0.134 for degree of smile. 

\begin{table}
\begin{center}
\caption{Comparison of inception score (IS) on Deep Fashion Dataset. Quality of Full-body images generated by DGAN outperforms state-of-art methods.}
\label{tab:DF_IS}
\begin{tabular}{c|c}
\hline
Method & Inception Score (IS)\\
\hline
PG2 (G1+D) \cite{ma2017pose} & 3.091 \\ \hline
pix2pix \cite{isola2017image} & 2.640 \\ \hline
Varational U-Net \cite{esser2018variational} & 3.087 \\ \hline
PGDG \cite{sun2019pose} & 3.006 \\ \hline
DGAN & \textbf{3.301} \\ 
\hline
\end{tabular}
\end{center}
\end{table}

\begin{table}
\begin{center}
\caption{Accuracy of conditional image generation with DGAN. On average DGAN achieves 86.2\% accuracy on attribute conditioning.}
\label{tab:DGAN_accuracy}
\begin{tabular}{c|c|c}
\hline
Dataset & Attribute & Accuracy  \\
\hline
\multirow{2}{*}{Full body} & clothing style & 0.907 \\ 
\cline{2-3}
& pose  & 0.871 \\ \hline
\multirow{3}{*}{celebA-HQ} & gender & 0.891 \\ 
\cline{2-3}
 & hair-color & 0.789 \\ 
 \cline{2-3}
 & smile & 0.134 (RMSE) \\  
\hline
\end{tabular}
\end{center}
\end{table}

\section{Discussion}
\label{Discussion}

\subsection{Latent space separation}
\label{sec:latent_separation}
As mentioned in section \ref{sec:introduction} \& \ref{Approach}, DGAN is based on the assumption that an image attribute is linearly separable in latent space. We randomly split the dataset into 80-20  for training and testing the \textit{latent-attribute} block classifiers. Test accuracies are reported in Table \ref{tab:latent_separation}. As we can observe from Table \ref{tab:latent_separation}, we achieve over 79\% accuracy on all of the discrete attributes, suggesting that each discrete attribute is linearly separable. For the continuous valued attribute - degree of smile, we achieve a low RMSE value of 0.173 suggesting that a linear regression line in latent space provides a good representation for smile.  

\begin{table}
\begin{center}
\caption{Test accuracy of linear classifiers/regressors in latent space.}
\label{tab:latent_separation}
\begin{tabular}{c|c|c}
\hline
Dataset & Attribute & Accuracy  \\
\hline
\multirow{2}{*}{Full body} & clothing style & 0.812 \\ 
\cline{2-3}
& pose  & 0.79 \\  \hline
\multirow{3}{*}{celebA-HQ} & gender & 0.865 \\ 
\cline{2-3}
 & hair-color & 0.824 \\ 
 \cline{2-3}
 & smile &  0.173 (RMSE) \\ \hline
\end{tabular}
\end{center}
\end{table}

\subsection{Latent vector modification}

The modified latent vector $\boldsymbol{z'}$ obtained in equation \ref{eq:final_update} will always lie in the correct subspace of desired attribute values if the attributes are truly disentangled i.e. the separating hyperplanes of various attributes are orthogonal to each other. However, as we can observe from the cosine similarity table  (\ref{tab:corr_full_body} and \ref{tab:corr_celeba}), different attributes are not necessarily orthogonal. We empirically show that our approach results in moving the latent vector to the correct subspace with an average accuracy of 90\%. Table \ref{tab:latent_modification} shows the accuracy of our latent modification approach. Since the intersecting half-spaces form a convex polygon, future works could explore obtaining $\boldsymbol{z'}$ using convex optimization techniques.

\begin{table}
\begin{center}
\caption{Cosine similarity between hyperplanes of various attributes for Full body dataset. We see that pose and clothing style are not necessarily orthogonal. }
\label{tab:corr_full_body}
\begin{tabular}{c|c|c}
\hline
 & pose & clothing style  \\ \hline
pose & 1.0 & 0.57 \\ \hline
clothing style & 0.57 & 1.0 \\ \hline
\end{tabular}
\end{center}
\end{table}

\begin{table}
\begin{center}
\caption{Cosine similarity between hyperplanes of various attributes for CelebA-HQ dataset. Here black, brown and blond are one vs all classifiers for hair-color. Certain attribute values such as gender and black hair color are not orthogonal, while gender and smile are fairly disentangled. }
\label{tab:corr_celeba}
\begin{tabular}{c|c|c|c|c|c}
\hline
 & gender & black & brown & blond &  smile \\ \hline
gender & 1.0 & 0.51 & 0.20 & -0.48 & -0.11 \\ \hline
black & 0.51 & 1.0 & -0.19 & -0.91 & -0.33  \\ \hline
brown & 0.20 & -0.19 & 1.0 & -0.17 & 0.041  \\ \hline
blond & -0.48 & -0.91 & -0.17 & 1.0 & 0.31  \\ \hline
smile & -0.11 & -0.33 & 0.041 & 0.31 & 1.0  \\ \hline
\end{tabular}
\end{center}
\end{table}

\begin{table}
\begin{center}
\caption{Accuracy of our latent modification approach outlined in equation \ref{eq:final_update}. On average, each discrete attribute has an accuracy of about 90\% and continuous attribute has a low RMSE value. }
\label{tab:latent_modification}
\begin{tabular}{c|c|c}
\hline
Dataset & Attribute & Accuracy  \\
\hline
\multirow{2}{*}{Full body} & clothing style & 0.92 \\ 
\cline{2-3}
& pose  & 0.886 \\  \hline
\multirow{3}{*}{celebA-HQ} & gender & 0.91 \\ 
\cline{2-3}
 & hair-color & 0.87 \\ 
 \cline{2-3}
 & smile & 0.08 (RMSE) \\ \hline
\end{tabular}
\end{center}
\end{table}

\subsection{Disentanglement}
As noted by authors in StyleGAN \cite{karras2019style}, features in latent space are disentangled where $\boldsymbol{w}$ offers higher degree of disentanglement than $\boldsymbol{z}$. We leveraged this observation of feature-space disentanglement for conditioning as we outlined above. However, we chose to employ $\boldsymbol{z}$ for our experimentation since $\boldsymbol{w}$ is not randomly chosen, rather is the output of a feedforward deep neural network and hence does not span the entire $R^{512}$. Moreover, the subspace spanned by $\boldsymbol{w}$ is unknown and complex; it is not straight forward to formalize the move from one feature value subspace to another.

\section{Conclusion}
In this work we propose Directional GAN (DGAN), which utilizes \textit{StyleGAN} to not only generate high-resolution full-body human images but also condition on certain image attributes. Specifically, we show how one can vary binary, multi-class and continuous valued attributes and condition on one or more of such attributes in a single step. Moreover our approach is flexible enough to work with any generative models that have sufficiently disentangled image features in latent space. Our approach also works well with any choice of classifiers/regressors in the \textit{image-attribute} block. While DGAN allows for a great degree of control over attributes in the generation process by employing directional vectors in the latent space, the generated images have slight unintended variations in other attributes, which is likely due to entanglement of attributes in the latent space. Future work could explore moving the latent vector to the desired subspace by using convex optimization which doesn't necessarily require image attributes to be disentangled. 

DGAN could be extended to allow for modification of given real images. One could first invert the given real image back to latent vector using an optimization based inversion approach \cite{lipton2017precise}, an encoder based approach \cite{donahue2016adversarial} or a hybrid of both.
This latent vector then could be modified appropriately using DGAN.



{\small
\bibliographystyle{ieee_fullname}
\bibliography{refs}
}

\end{document}